# Leveraging Multi-domain, Heterogeneous Data using Deep Multitask Learning for Hate Speech Detection

**Prashant Kapil**
Department of CSE
Indian Institute of Technology Patna
prashant.pcs17@iitp.ac.in

**Asif Ekbal**
Department of CSE
Indian Institute of Technology Patna
asif@iitp.ac.in

## Abstract

With the exponential rise in user-generated web content on social media, the proliferation of abusive languages towards an individual or a group across the different sections of the internet is also rapidly increasing. It is very challenging for human moderators to identify the offensive contents and filter those out. Deep neural networks have shown promise with reasonable accuracy for hate speech detection and allied applications. However, the classifiers are heavily dependent on the size and quality of the training data. Such a high-quality large data set is not easy to obtain. Moreover, the existing data sets that have emerged in recent times are not created following the same annotation guidelines and are often concerned with different types and sub-types related to hate. To solve this data sparsity problem, and to obtain more global representative features, we propose a Convolution Neural Network (CNN) based multi-task learning models (MTLs)[1] to leverage information from multiple sources. Empirical analysis performed on three benchmark datasets shows the efficacy of the proposed approach with the significant improvement in accuracy and F-score to obtain state-of-the-art performance with respect to the existing systems.

## 1 Introduction

The continuing rise in the usage of the internet has loaded a large volume of content in social media like Twitter with 500 million[2] tweets per day and Facebook laden with 510K comments[3] per minute. These sites are important sources for people to give their opinion on a host of general social topics. The message can be clean, sarcastic, obscene, offensive, rude, hateful, etc. (Nockleby, 2000) defined *hate speech* as a broad umbrella term that describes it as any communication that demeans any person or any group based on race, color, gender, ethnicity, sexual orientation, and nationality.

Defining hate is, itself, a difficult task as it greatly depends on the demography, i.e. the same content comes under *Right to speech* in some countries, while other countries might adhere to a very strict policy for the same message.

In recent times, Germany made policy for the social media companies that they would have to face a penalty up to $60 million[4] if they failed to remove illegal content on time. Denmark and Canada have laws that prohibit all the speeches that contain insulting or abusive content that could promote violence and social disorders. The Indian government has also urged leading social media sites such as Facebook, Twitter to take necessary action against hate speech, especially those posts that create social outrage. Setting aside legal actions our aim should be to combat these texts by agreeing to a set of standard definitions, guidelines, and practices to remove the content. Recently many automated techniques following supervised learning utilizing deep neural networks have been developed. Recently shared tasks such as (Basile et al., 2019; Mandl et al., 2019; Zampieri et al., 2019) have mainly focused on developing multiple-layer identification of offensive languages. The existing prior research towards this direction mainly focused on single-task learning (STL) where classification task on one data set at a time is solved by training the model in stochastic gradient descent approach. However, training of neural networks relies on a large amount of data, and creating a balanced data set seems to be time-consuming, and tedious. As the number of posts showing aggressive tendencies is very less

---
[1] code is available at https://github.com/imprasshant/STL-MTL
[2] https://www.internetlivestats.com/twitter-statistics/
[3] https://kinsta.com/blog/facebook-statistics/
[4] https://www.inc.com/joseph-steinberg/germanys-tough-new-social-media-law-punishes-offensive-posts-with-fines-of-up-to-60-million.html

compared to non-aggressive posts, we leverage the concept of homogeneous multi-task learning where we utilized the multiple classification task data sets to be trained jointly to solve the task. Although binary classification is very problematic, it filters out harmful messages and provides hateful data to further train the model to classify the data into more fine-grained classes and helps in getting the target and sentiment behind the posts, thus preventing the violation of the right to freedom of speech.

The key characteristics of our current work are summarized as follows.

(i). We propose a deep multi-task learning framework that leverages information from multiple sources. We experiment on five different variations of CNN based single-task learning (STL) and five different variations of CNN based multi-task Learning (MTL) approaches for solving the problem of hate speech classification.

(ii). The proposed classification approach can be utilized to obtain hateful or abusive posts to further train any classifier with these data to perform the classification to finer labels.

## 2 Related Work

In recent times, online hate speech detection has attracted the attention of researchers and developers because of its necessity in maintaining social fabrics. In recent times, most of the methods that have emerged are mainly based on classical machine learning and deep learning. (Badjatiya et al., 2017) defined hateful tweets as speech that contains abusive speech targeting *individuals* (e.g. cyberbullying, politician, a celebrity or a product) or a *group* (a religious group, country, LGBT, gender, organization), etc. (Wulczyn et al., 2017) identified the personal attack as a binary classification problem and experimented with logistic regression and multilayer perceptron with word n-grams or char n-grams based features. (Nobata et al., 2016) observed that including simple n-gram features are more powerful than linguistics-based features, syntactic-based features, as well as word and paragraph embeddings. (Davidson et al., 2017) highlighted the differences between hateful and offensive languages and that, conflating these two erroneously will make many speakers be hate speakers. They highlighted the need to train the model with hateful data that does not contain any particular keywords or abusive terms to enrich the model with more contextual and knowledge-based features. (Chakrabarty et al., 2019) provided visualization of attention weights and concluded that the model assigned higher attention to potentially abusive terms when employed with contextual information in comparison to self-attention based features. (MacAvaney et al., 2019) utilized BERT, that make use of Transformer (Vaswani et al., 2017), an attention mechanism helping to capture the contextual representation between words and sub-words of a sentence that is utilized to perform the classification task on (de Gibert et al., 2018). (Pérez and Luque, 2019) leveraged BiLSTM with a dense layer on top consuming Elmo vectors (Peters et al., 2018), and Bag of words as additional input to do the classification on the data by (Basile et al., 2019). In this paper, we present a multi-task framework that aims at leveraging information contained in multiple related tasks and improve the classification performance of the hate data sets.

## 3 Methodology

### 3.1 Preprocessing

We perform different steps of pre-processing to clean the text.

1. A light pre-processing by removing all the characters like @!:;?. and removing all the numbers (0-9), URLs present in the tweet.

2. Word segmentation is being done to convert the hashtags like *#BuildTheWall* → build the wall, *#SendthemBack*→ send them back, *#refugeeswelcome* → refugees welcome, *#humantrafficking* → human trafficking, *#whitegenocides* → white genocides, *#makeLoveNotWar* → make love not war, *#F\*\*kracism*→ f\*\*k racism, etc. using python (Rossum, 1995) *word segment* to preserve the important features to compute sentiment of any type of message.

3. All the emoticons were manually categorized into 5 categories, i.e. *love, sad, happy, shocking* and *anger*. The unicode character of emoticons is then substituted with the token it matched.

4. All the @ mentions were replaced with the common token i.e *user*.

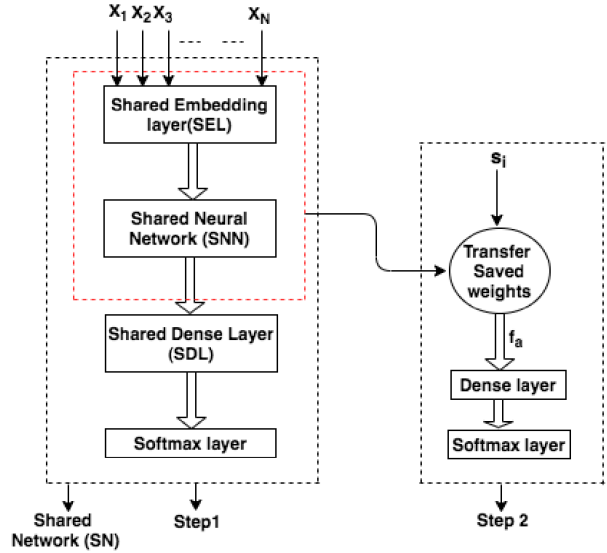

Figure 1: **Architecture of Fully Shared Multi-Task Learning(FS-MTL)**

### 3.2 Embeddings

*Word embeddings* ($w_e$): In our experiments, we utilized the Google pre-trained *word2vec* vectors trained on 100 billion words to produce 300 dimensions for each word capturing the semantic and syntactic relationship between the words trained using skip-gram by (Mikolov et al., 2013).

*Character embedding* ($c_e$): The presence of Out-of-Vocabulary (OOV) word is a serious problem in a social media text. Embedding for such words in the pre-trained word embedding model is not found, hence losing morphological information. We leverage the skip-gram model by (Bojanowski et al., 2017) which represents each word as a bag of character n-grams. The dimension of each word using character embedding is 300. The final word embedding $x_e$ for word x $\in$ X is represented by the following process:

$$x_e = w_e \oplus c_e \tag{1}$$

where ($\oplus$) denotes the concatenation operation and X is the number of unique tokens. The resulting dimension of $x_e$ is 600.

### 3.3 Models

We adopt (Kim, 2014) for word-level Convolutional neural networks(*Word-CNN*) and (Zhang et al., 2015) for Character-level Convolutional neural networks (*Char-CNN*) to build different variants of CNN for our experiments. Here, we describe Word-CNN and Char-CNN.

**Word-CNN**: We adopted the CNN-Static by (Kim, 2014). The input sequence $S_i$ of length $l$ is tokenized to assign a unique integer index to each word $w_i$, that is then mapped to its $N$ dimension real-valued vector. A *convolution* operation involves a filter $f \in R^{hN}$ which is applied to the $h$ words to produce a new feature $x_i$ in Eq.2. Here, $b \in R$ is a bias term and $g$ is a non-linear activation function. This process is repeated *l-h+1* to get the feature map $x$ in Eq.3.

$$x_i = g(f \cdot S_{i:i+h-1} + b) \tag{2}$$

$$x = [x_1, x_2, ......, x_{l-h+1}] \tag{3}$$

Then the pooling layer is applied to reduce the spatial size of the representation helping in reducing over-fitting. The vector form of features obtained from the last CNN layer is fed into the fully connected layer followed by the softmax activation function that calculates the probability values for all the classes. We define the following 5 models that utilize the word-level CNN: *Model 1*[5], *Model 2*, *Model 4*, *Model 5*, *Model 6 Model 7*, *Model 8*, *Model 9* and *Model 10*(all these models are defined below).

**Char-CNN**: We adopt the character-CNN (Zhang et al., 2015) where the gradients are obtained by backpropagation to perform optimization. It accepts a sequence of encoded characters as input. The encoding is done by quantizing each character using 1-of-*m* encoding, also known as *one-hot-encoding*. Then the sequence of characters is transformed to a sequence of *m* sized vector with fixed length $l_o$= 256/1024. The value of *m* in their proposed model is 70 with 26 for the English alphabet, 10 digits, 33 other characters, and one for the newline character. They designed 9 layers with 6 convolutional layers and 3 fully-connected layers. They initialized the weights using Gaussian distribution. Two models, *viz. Model 3* and *Model 4* (all these models are defined below) utilize the concept of character CNN. Below, we briefly describe each of the proposed models.

---

[5]Model *i* and model *i* will refer to the same model in the text where $i \in [1, 10]$

**Model 1:** *Random word vectors-CNN*: We adopt the method by (Gambäck and Sikdar, 2017) to assign the random vector of dimension 600 as feature embeddings for words.

**Model 2:** *Word-CNN*: In this model, we utilize real-valued vectors of 600 dimensions for each word capturing the semantic and syntactic relationship between the words from Eq.1.

**Model 3:** *Char-CNN*: Our designed model consists of representing each characters using 27 sized vector with 26 elements for the English alphabet and one for all other symbols. This model consists of a convolution layer with kernel size 4 followed by max-pool layer of size 3. This is fed into another convolution layer with kernel size 4 and max-pool layer of size 3. This is followed by 2 dense layers of size 64 and 2. The strides used in convolution layers are 4 and 2.

**Model 4:** *Hybrid-CNN*: It utilizes both character and word input at the same time. The output of both the channels after flattening the pooling features is concatenated to pass into a fully connected layer with softmax activation function.

**Model 5:** *CNN-Word-Attention:* This mechanism expands the functionality of neural networks by paying attention to the specific parts of the sentence depicting the human brain. We utilized the CNN-sentence-level attention by (Raffel and Ellis, 2015). It calculates the attention weight for the important words to form a representation of the sentences. Each word's hidden state representation ($h_t$) is passed through a learnable function a($h_t$) to produce probability value $\alpha_1, \alpha_1...\alpha_n$ for each word. The sentence vector *output* is calculated by the weighted average of $h_t$ with weights of $\alpha$.

$$e_t = tanh(Wh_t + b) \quad (4)$$
$$\alpha_t = softmax(e_t) \quad (5)$$
$$output = \sum_{t=1}^{t=n} \alpha_t h_t \quad (6)$$

**Fully Shared MTL(FS-MTL):** The architecture of model 6, model 7 and model 8 are based on this scheme that is shown in Figure 1. This scheme consists of two steps.

**Step 1: Training of Shared Network (SN):** The SN consists of 4 components: Shared Embedding Layer (SEL), Shared Neural Network (SNN), Shared Dense Layer (SDL) and Softmax layer. This network is pre-trained by taking equal samples from each of the participating data sets and training it in batch-wise manner. The Shared Embedding Layer (SEL) consists of the unique tokens from all the data sets. All the different classes of each data set are merged to represent class $c_i$ where in this experiment i $\in$ [1,N] where N is number of data set. The parameters of the SN are trained to minimize the *categorical cross entropy* of the predicted and true distribution on all the tasks. The loss $L_{Task}$ can be defined as:

$$L_{Task} = \sum_{k=1}^{K} \alpha_k \cdot L(\hat{y}^k, y^k) \quad (7)$$

where $\alpha_k$ is the class weight i.e 1 in this experiment and L($\hat{y}$,y) is defined in equation 8.

$$L(\hat{y}, y) = -\sum_{i=1}^{C}\sum_{j=1}^{N} y_i^j \log \hat{y}_i^j \quad (8)$$

Here C is the total number of classes, N is the number of samples, $y_i^j$ is the ground truth label and $\hat{y}_i^j$ is the predicted label.

**Step 2**: The trained shared network (SN) is sliced off to extract the weight matrix of the first two layers: SEL and SNN, denoted in red color in Figure 1[6]. The parameters of the transferred layers to the new network are kept frozen. A new sentence $S_i$ is passed through the frozen weight matrix to get representation $f_a$ which is passed to dense layer followed by softmax layer to get the probability values.

**Model 6:** *Word-CNN-Fully Shared MTL*: We adopt the schema of (Liu et al., 2017) by

---
[6]The figure is best viewed in color

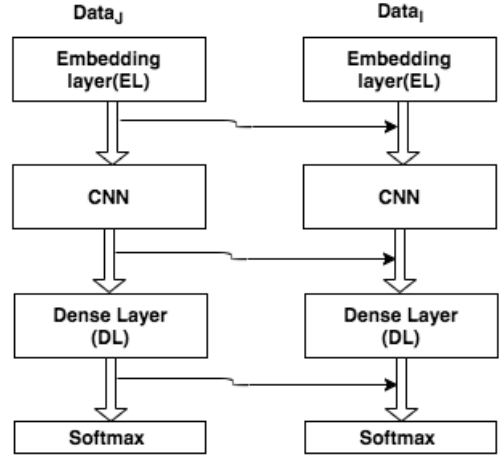

Figure 2: **Architecture of Soft Sharing MTL**

employing fully-shared Word-CNN layers to extract features for all the tasks. It takes the view that features of *task m* can be totally shared by *task n* and the vice-versa. Figure 1 explains the idea. Here $X_1 \to D_1$, $X_2 \to D_2$, $X_3 \to D_3$. $\{D_1, D_2$ and $D_3$ explained in section 4$\}$

**Model 7**: This model utilizes the sentiment data to be trained with hate data in a fully shared manner. Here $X_1 \to D_i$, $X_2 \to S_1$, $X_3 \to S_2$, $X_4 \to S_3$.
$\{S_1, S_2$ and $S_3$ explained in section 4$\}$ and i = 1,2,3

**Model 8**: The intermediate feature $f_a$ obtained from model 6 and model 7 is concatenated to pass into dense layer followed by softmax layer.

**Model 9**: *Soft Sharing CNN-Word-MTL:* This model is motivated by (Xiao et al., 2018) that utilizes the CNN based multitasking paradigm. Every task owns a subnet and shares the features with each other. The embedding layer (EL) in Figure 2 consists of uniques tokens present in all the data sets. Here $D_1$, $D_2$ and $D_3$ will share feature with each other. All the subnet undergoes a pre-training of the text sequences. Let C be the total collection of *n* tasks C = $\{T_1, T_2, ..., T_n\}$. The output of any sequence $s_i$ at any layer $l$ is the concatenation of the output of the same sequence $s_i$ from all the other tasks. Task *i* borrows the features from Task *j* which is calculated as

$$g_{ij}^l = (W_{ij}^l \cdot F_{ij}^l + b_{ij}^l) \quad (9)$$

where l denotes the level of layers. For any task i the output of $F_{ij}^{l+1}$ by merging the $F^l$ from all the other tasks by

$$F_{ij}^{l+1} = \sum_{j \in C, j \neq i} g_{ij}^l + F_i^l \quad (10)$$

**Model 10**: The training will remain the same as of model 9 but $D_i$ will share the features with $S_1$, $S_2$ and $S_3$. Figure 2 explains the idea for 2 task which can be extended to *n* tasks. Here i $\in$ [1,3]

## 4 Dataset and Experiment Setup

### 4.1 Data

We evaluate our model on 3 different datasets. (denoted as $D_1$, $D_2$ and $D_3$). Hate speech and sentiment analysis are closely related, and it is safe to assume that usually negative sentiment pertains to a hate speech message. We also utilized 3 sentiment data which have been described as $S_1$, $S_2$ and $S_3$. Table 1 and Table 2 shows the statistics of the datsets.

$D_1$(de Gibert et al., 2018): The sentences have been extracted from *stormfront*, a white supremacist forum. A subset of 22 sub-forums covering diverse topics and nationalities was random-sampled to gather individual posts uniformly distributed among the sub-forums and users. The most common hateful words found were *ape, homosexuals, libtard, monkey* and *miglet*. The data set constitutes of 36.05% and 41.63% hate vocabulary from *gender* and *ethnicity*.

$D_2$(Basile et al., 2019): This dataset is part of hate speech against immigrants and women in English, collected between July to September 2018. The most frequent keywords were *migrant*,

*refugee, b\*\*ch, #buildthatwall, h\*e* and *women*.

$D_3$(Mandl et al., 2019): The HASOC dataset was subsequently sampled from Twitter and partially from Facebook for the three languages. For our experiments, we leveraged the English data. They identified topics for which many hate posts can be expected. Thus, the tweets were acquired using hashtags and keywords that contained offensive contents.

$S_1$[7]: This dataset is crawled from twitter containing US Airline Sentiment tweets.

$S_2$(Rosenthal et al., 2017): The English topics based on popular current events that were trending on Twitter, both internationally and in specific English speaking countries were used to crawl using Twitter API. The topics included a range of named entities (e.g., Donald Trump, iPhone), geopolitical entities (e.g., Aleppo, Palestine), and other entities (e.g., Syrian refugees, Dakota Access Pipeline, Western media, gun control, and vegetarianism).

$S_3$(Misra and Arora, 2019): To overcome the limitations related to noise in Twitter datasets, they collected a news Headlines dataset from two news websites. The Onion2 aims at producing sarcastic versions of current events and they collected all the headlines from News in Brief and News in Photos categories (which are sarcastic). They collect real (and non-sarcastic) news headlines from HuffPost.

Table 1: **Hateful Data Statistics**

| Dataset | labels and count | Test |
|---|---|---|
| $D_1$ | Hate:1097 <br> Non-Hate:8571 | CV |
| $D_2$ | Hate:4210 <br> Non-Hate:5790 | Hate:1260 <br> Non-Hate:1740 |
| $D_3$ | Hate:2261 <br> Non-Hate:3591 | Hate:288 <br> Non-Hate:865 |

### 4.2 Experimental Setup

All the deep learning models were implemented using Keras, a neural network API by (Chollet et al., 2018) with Tensorflow (Abadi et al., 2016) at the backend. All the dataset were split using 5-fold cross validation mode using the Stratified K-fold with 80% for training and 20% for testing with an equal proportion of samples from all the classes. The batch size of 30 is used for training SN Model in Figure 1. The official test set of 3000 instances is utilized for $D_2$ and 1153 for $D_3$. Random search was done to fine-tune the results of the neural networks to select the best performing combination of hyperparameters. Categorical cross-entropy is used as loss function with *adam*, a combination of *Adagrad* and *RmsProp* is used as the optimizer. The number of filters used for Word-CNN and Char-CNN are 100 and 256. The value for *bias* is randomly initialized to all zeros, *relu* activation function were employed at the intermediate layer and *softmax* is utilized at last dense layer.

---

[7] https://www.kaggle.com/crowdflower/twitter-airline-sentiment

Table 2: **Sentiment Data Statistics**

| Dataset | labels and count | Test |
|---|---|---|
| $S_1$ | Positive:2363 <br> Negative:9178 <br> Neutral:3099 | - |
| $S_2$ | Positive:7059 <br> Negative:3231 <br> Neutral:10342 | - |
| $S_3$ | Sarcastic:25358 <br> Non-sarcastic:29970 | - |

## 5 Results, Comparison and Analysis

We report the 5-Fold cross validation result for $D_1$, $D_2$ and $D_3$ in Table 5, Table 6 and Table 7. Of the 10 models all the 5 CNN-MTL[8] outperforms 5 CNN-STL based approaches. We report here the macro-f, weighted-f and accuracy of the proposed methods. The best models for STL in $D_1$, $D_2$ and $D_3$ are CNN-attention, character-CNN and CNN-attention respectively. Hybrid-CNN also shows significant improvements in $D_2$. For the MTL based approach, concatenation of sentiment and hate based features gives good results for $D_1$. For $D_2$ and $D_3$ model 7 and model 6 performs best. Table 8, Table 9 and Table 10 enlist the comparisons between the previous benchmarks with our best models.

---

[8] The results obtained for $S_1$, $S_2$ and $S_3$ in MTL setting is not reported to focus only on hateful text detection.

Table 3: **False Negatives**

| Sl | Type | Sentence |
|---|---|---|
| 1. | Toxic | B**ch our streak is dying @C**t***Lady |
| 2 | Toxic | @AMike4761 Wake the f**k up and fight back! Savethewest sendthemback |
| 3 | Non-Toxic | Fed up with this crap! #DeportThemAll |
| 4 | Non-Toxic | @globalnews #SendThemHome we do not need those #Students here in #Canada |
| 5 | Direct Attack | "The first sexual attack against a woman happened in Hungary. The criminal is a ""legal"" afghan rapefugee. This is how Orban protects us." |
| 6 | Direct Attack | @whaas3 @judithineuropa Just got on twitter because of this farce today. Imagine this, I make a report on You. Calling you names and telling people how big liar and a****le you are without reason. Would you be angry |
| 7 | Doubtful | @DVATW @TheHairyJobbie By god there is a lot of woman and children got off that boat eh,poor holiday makers paying good money to go on holiday to witness that roundthemup sendthemback |
| 8 | Doubtful | @AdamBandt if its gets you upset - Hes the best man for the job. #gohome #strongborders #sendthemback |
| 9 | Sarcastic | "Please don't call it ""rescue"" - it's human trafficking #portsclosed #sendthemback #benefitseekers" |

Table 4: **False Positives**

| Sl | Type | Sentence |
|---|---|---|
| 1 | Toxic | H*e stood behind a car door and said "I don't feel comfortably with you that close" b**ch made |
| 2· | Toxic | @Cornjdw Lmao f**k you bitch don't get mad at me cuz u don't know the game of basketball hoe |
| 3· | Non-Toxic | A little louder @w_terrence for the liberals in the back. #SendThemBack #BuildTheWall |
| 4 | Non-Toxic | Meanwhile in Spain..stopimmigration |
| 5 | Direct Attack | @realDonaldTrump Do you support @realDonaldTrump's Southern Border Wall? Vote #RETWEETCheck out wall progress at Order Bricks to show you. |
| 6 | Direct Attack | Still can't be. Even the neonazis behind the ""rapefugee"" website only claim around 450 rapes by immigrants for all of 2016. |
| 7. | References | What an idiot. #buildthatwall |
| 8. | References | @RepLowenthal Asylum seekers should enter at a LEGAL #USA port of entry |

Table 5: **Evaluation Results on $D_1$ (de Gibert et al., 2018)**

| Model | Macro(%) | Weighted(%) | Acc.(%) |
|---|---|---|---|
| Single Task Learning | | | |
| **Model-1** | 47.44 | 83.30 | 88.39 |
| **Model-2** | 47.25 | 83.36 | 88.61 |
| **Model-3** | **48.48** | 83.13 | 87.47 |
| **Model-4** | 47.07 | 83.33 | 88.63 |
| **Model-5** | 47.79 | **83.49** | **88.65** |
| Multi Task Learning | | | |
| **Model-6** | 87.81 | 95.13 | 95.18 |
| **Model-7** | 85.19 | 93.93 | 93.84 |
| **Model-8** | **90.55** | **96.35** | **96.52** |
| **Model-9** | 84.17 | 93.70 | 93.78 |
| **Model-10** | 72.11 | 89.76 | 90.81 |

Table 6: **Evaluation Results on $D_2$ (Basile et al., 2019)**

| Model | Macro(%) | Weighted(%) | Acc.(%) |
|---|---|---|---|
| Single Task Learning | | | |
| **Model-1** | 46.69 | 44.33 | 50.76 |
| **Model-2** | 48.04 | 45.91 | 51.36 |
| **Model-3** | **51.49** | **51.19** | 51.56 |
| **Model-4** | 49.78 | 47.66 | **52.76** |
| **Model-5** | 45.47 | 42.76 | 50 |
| Multi Task Learning | | | |
| **Model-6** | 91.40 | 91.57 | 91.54 |
| **Model-7** | **93.60** | **93.75** | **93.75** |
| **Model-8** | 93.41 | 93.56 | 93.56 |
| **Model-9** | 90.11 | 90.35 | 90.35 |
| **Model-10** | 89.43 | 89.72 | 89.76 |

## 5.1 Error Analysis

**Quantitative Analysis:** The confusion matrix obtained by best performing model on $D_1$, $D_2$ and $D_3$ is presented in Table 11, Table 12 and Table 13. For $D_1$ model 8 performs best, model 7 is performing best for $D_2$ and $D_3$. From the table it can be seen that misclassification rate in the proposed model for *hate* is 23% for $D_1$, 31.5% for $D_2$ and 34% for $D_3$. However, the misclassification for *Non-Hate* is 1.6% for $D_1$, 54.25% for $D_2$ and 13.87% for $D_3$.

**Qualitative Analysis:** We also identified some of the false negative cases i.e hateful tweet predicted to non-hate and false positive cases i.e non-hate tweet classified as hateful class in Table 3

Table 7: **Evaluation Results on $D_3$ (Mandl et al., 2019)**

| Model | Macro(%) | Weighted(%) | Acc.(%) |
|---|---|---|---|
| **Single Task Learning** | | | |
| Model-1 | 57.32 | 61.47 | 65.12 |
| Model-2 | 56.98 | 61.08 | 64.54 |
| Model-3 | 44.27 | 51.37 | 61.77 |
| Model-4 | 59.82 | 63.14 | 65.12 |
| Model-5 | **62.19** | **65.27** | **67.05** |
| **Multi Task Learning** | | | |
| Model-6 | **87.37** | **87.99** | **87.96** |
| Model-7 | 79.82 | 81.20 | 81.65 |
| Model-8 | 86.76 | 87.65 | 87.93 |
| Model-9 | 86.17 | 86.94 | 87.01 |
| Model-10 | 84.34 | 85.13 | 85.12 |

Table 8: **Comparison to the state-of-the-art systems and proposed system for $D_1$ (de Gibert et al., 2018)**

| Model | Macro(%) | Acc(%) |
|---|---|---|
| (MacAvaney et al., 2019) | 82.01 | 82.01 |
| (MacAvaney et al., 2019) | 80.31 | 80.33 |
| (Berglind et al., 2019) | 70.80 | 72.20 |
| (Berglind et al., 2019) | 81.90 | 81.60 |
| (Berglind et al., 2019) | 78.40 | 77.10 |
| Model-8 | **90.55** | **96.52** |
| Model-6 | 87.81 | 95.18 |

Table 9: **Comparison to the state-of-the-art systems and proposed system for $D_2$ (Basile et al., 2019)**

| Model | Test Data | |
|---|---|---|
| | Macro(%) | Acc(%) |
| (Ding et al., 2019) | 54.60 | **56** |
| (Montejo-Ráez et al., 2019) | 51.90 | - |
| (Pérez and Luque, 2019) | 47.10 | 50.80 |
| (Baruah et al., 2019a) | 51 | 54 |
| Model-7 | **55.24** | 55.26 |
| Model-8 | 50.55 | 52.60 |

Table 10: **Comparison to the state-of-the-art systems and proposed system for $D_3$ (Mandl et al., 2019)**

| Model | Test Data | |
|---|---|---|
| | Macro(%) | Acc(%) |
| (Mishra and Mishra, 2019) | 74.65 | - |
| (Baruah et al., 2019b) | 74.62 | - |
| (Jiang, 2019) | 74.31 | - |
| Model-7 | **75.39** | **81.09** |
| Model-8 | 74.68 | 80.65 |

Table 11: **Confusion matrix of $D_1$ (de Gibert et al., 2018)**

| Class | Hate | Non-Hate |
|---|---|---|
| Hate | 844 | 253 |
| Non-Hate | 136 | 8435 |

and Table 4. It points to the fact that due to usage of words like f**k, bi**h, kill, bas***d etc. in both hate and non-hate context, the neural network

Table 12: **Confusion matrix of $D_2$ (Basile et al., 2019)**

| Class | Hate | Non-Hate |
|---|---|---|
| Hate | 862 | 398 |
| Non-Hate | 944 | 796 |

Table 13: **Confusion matrix of $D_3$ (Mandl et al., 2019)**

| Class | Hate | Non-Hate |
|---|---|---|
| Hate | 190 | 98 |
| Non-Hate | 120 | 745 |

is being confused to classify correctly.

## 6 Conclusion and Future work

In this paper we have proposed five multi-task learning based approaches for hate speech detection. The proposed approaches has an ability to learn shared features between three different hate speech data sets and also leveraging the knowledge from the data of sentiment analysis tasks. The efficacy of the proposed approach is evident from the fact that it shows a consistent improvement in the F-score and accuracy values over the models working on single-task learning paradigm.

The system failure on some cases highlights the need to build a more diverse and robust neural network system to take into account the contextual, demographic as well as the knowledge based features.